\documentclass{article}
\usepackage[a4paper, total={6in, 9.5in}]{geometry}
\usepackage[dvipsnames]{xcolor}
\usepackage{mathtools} \usepackage{multirow}
\usepackage{booktabs}
\usepackage{colortbl}
\usepackage{subfigure}
\usepackage[export]{adjustbox}
\usepackage{afterpage}

\usepackage{wrapfig}

\usepackage{caption}

\usepackage[utf8]{inputenc} 
\usepackage[T1]{fontenc}    
\usepackage{hyperref}       
\usepackage{url}            
\usepackage{booktabs}       
\usepackage{amsfonts}       
\usepackage{nicefrac}       
\usepackage{microtype}      
\usepackage{xcolor}         

\def\eg{\emph{e.g.~}}

\def\ie{\emph{i.e.~}}

\usepackage{tikz}

\title{Logits are predictive of network type}

\author{
Ali Borji \\
Quintic AI, San Francisco, CA \\ 
\texttt{aliborji@gmail.com} 
}

\begin{document}

\maketitle

\begin{abstract}


We show that {\bf it is possible to predict which deep network has generated a given logit vector} with accuracy well above chance. We utilize a number of networks on a dataset, initialized with random weights or pretrained weights, as well as fine-tuned networks. A classifier is then trained on the logit vectors of the trained set of this dataset to map them to their corresponding network index. The classifier is then evaluated on the test set of the same dataset. Results are better with randomly initialized networks, but also generalize to pretrained networks as well as fine-tuned ones. Classification accuracy is higher using unnormalized logits than normalized ones. We find that there is little transfer when applying a classifier to the same networks but with different sets of weights. 
In addition to help better understand deep networks and the way they encode uncertainty, we anticipate our finding to be useful in some applications (\eg tailoring an adversarial attack to a certain type of network). Code is available at \href{https://github.com/aliborji/logits}{\underline{link}}.

\end{abstract}

\section{Motivation}
Deep neural networks are prevalent in many domains and achieve astonishing performance. Our understanding of their behavior, however, is still limited. Here, we present a new surprising phenomenon which is a bit hard to explain. We show that it is possible to predict the type of deep network from the logits that it generates. Exploring the reasons underlying this phenomenon helps better understand high dimensional representations that are learned by deep networks. Our finding suggests that perhaps deep networks have different signatures of uncertainty encoding. For some applications, it may be useful to know which network has generated the output, for example when a customized treatment is necessary for a task (\eg knowing which network has generated an adversarial attack for efficient defense against it).

\section{Method}

\subsection{Experimental setup}
We choose a dataset and a number of deep networks. We first feed the training set of this dataset to the networks and collect the logits (real-valued vector $s$). Next, we form the following input-output pairs $(s_{ij}, i)$ where indices $i$ and $j$ represent network $i$ and input image $j$, respectively. We then train a classifier to map the logits to the network index. The goal is to predict which network has generated a given logit vector $s$. The classifier trained on the training set of the dataset (here a SVM with linear or RBF kernels) is evaluated on the test data of the same dataset. 

Since the logits produced by different networks might have different ranges of values, due to network hyperparameters and internal dynamics, we experiment with both unnormalized and normalized (logits divided by their sum) logits. The PyTorch code to extract the logits is as follows:

\begin{verbatim}
    Unnormalized: outputs = model(inputs).cpu()
    Normalized: outputs = outputs/outputs.sum(axis=1)[:,None]
\end{verbatim}

Three scenarios are considered: 1) networks initialized with random weights, 2) networks initialized with pretrained ImageNet weights, and 3) fine-tuned networks. On some dataset, we consider training the networks from scratch. We also study the effect of the number of training data on network prediction accuracy. Additionally, we explore whether a classifier trained on logits of some networks can predict the network type from logits of the same networks but when they are initialized or trained differently (\ie generalization or transfer). 


\subsection{Deep networks}

We choose 9 state-of-the-art object recognition networks\footnote{Networks are available in PyTorch hub: \url{https://pytorch.org/hub/}. We used a 12 GB NVIDIA Tesla K80 GPU to conduct the experiments.}. These networks have been published over the past several years and have been immensely successful over the ImageNet benchmarks~\cite{deng2009imagenet}. They include
ResNet101 and ResNet152~\cite{resnet}, AlexNet~\cite{krizhevsky2012imagenet}, DenseNet~\cite{huang2017densely}, GoogleNet~\cite{szegedy2015going}, 
MobileNetV2~\cite{sandler2018mobilenetv2}, ResNext~\cite{xie2017aggregated}, ResNext\_WSL~\cite{mahajan2018exploring}, and Deit~\cite{touvron2021training}. The code to download the weights is as follows:

\begin{verbatim}
torch.hub.load(`pytorch/vision:v0.10.0', `resnet101', 
                pretrained=trained, force_reload=True, skip_validation=True)    
\end{verbatim}

The variable \texttt{trained} can be False or True, corresponding to Random or Pretrained weights.

\begin{figure}[t]       
    \centering 
    \includegraphics[width=.47\linewidth]{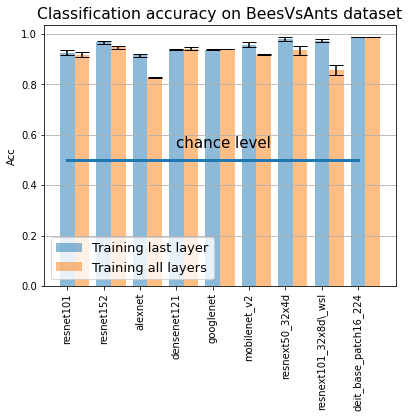}
    \includegraphics[width=.47\linewidth]{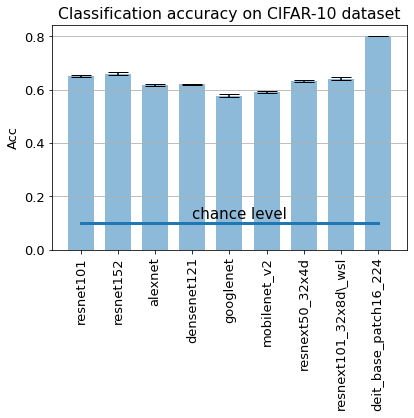}
    
    \caption{\textbf{Classification accuracy of deep networks}. Left) Binary classification over BeesVsAnts dataset (trained for 10 epochs). Right) 10-way classification over CIFAR-10 dataset (trained for 6 epochs; only the last layer is fine-tuned). Networks are initialized with ImageNet weights. Results are averaged over 3 runs.}
    \label{fig:accs}
\end{figure}

\section{Experiments and Results}
We conduct four experiments over three datasets including BeesVsAnts, CIFAR-10, and MNIST to validate our hypothesis. The networks over the former two datasets are the ones mentioned above. Over the MNIST dataset, we train simple CNNs with few layers of convolution and pooling. 

We fine-tune (or train) networks using Cross Entropy loss, SGD optimizer, and learning rate (LR) of 0.001, and momentum of 0.9. LR is decayed by a factor of 0.1 every 7 epochs. Networks are fine-tuned (or trained) for 10 epochs on BeesVsAnts and MNIST, and 6 epochs on CIFAR-10 dataset. The datasets to train and test the network prediction classifier are balanced, \ie the same number of logits per network type.


\subsection{Experiment I: Bees vs. Ants classification}
\begin{figure}[t]       
    \centering 
    \includegraphics[width=.47\linewidth]{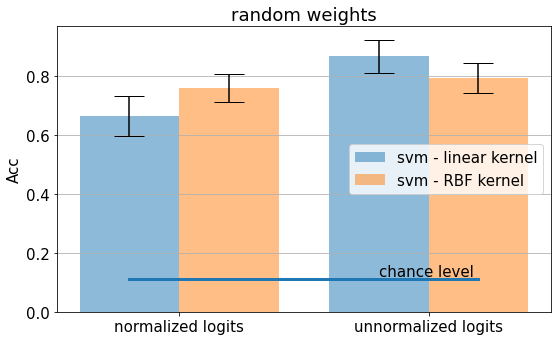}
    \includegraphics[width=.47\linewidth]{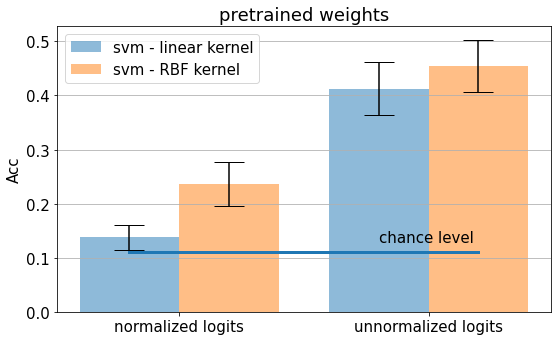}
    
    \includegraphics[width=.47\linewidth]{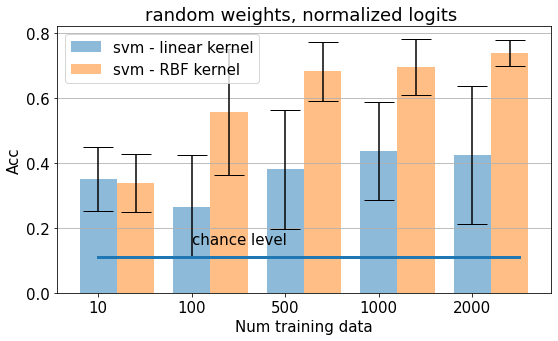}
    \includegraphics[width=.47\linewidth]{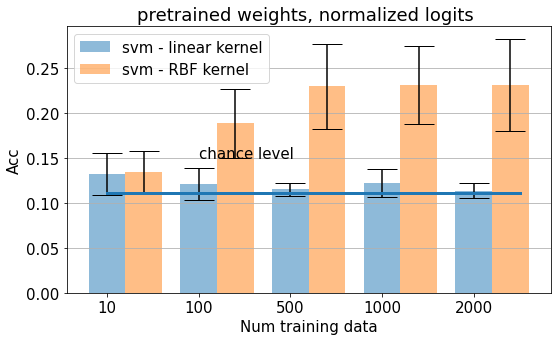}

    \includegraphics[width=.47\linewidth]{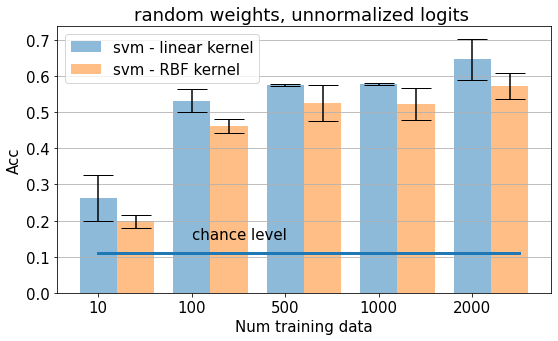}
    \includegraphics[width=.47\linewidth]{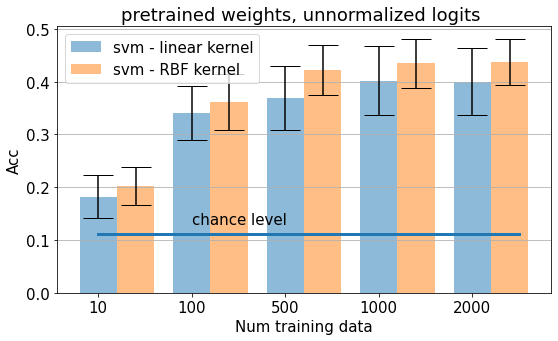}
    
    \includegraphics[width=.47\linewidth]{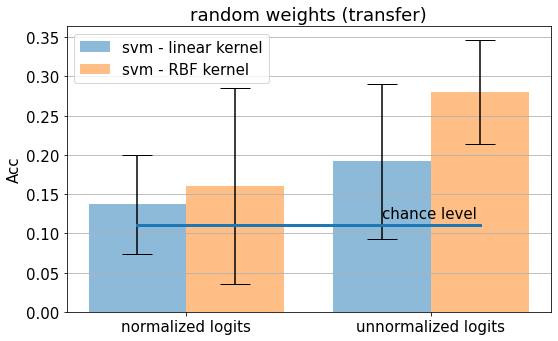}
    \includegraphics[width=.47\linewidth]{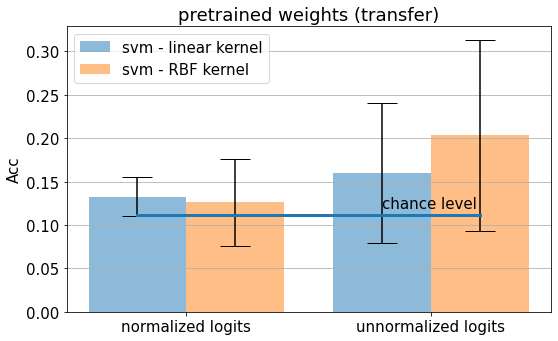}

    \caption{\textbf{Results over the BeesVsAnts dataset}. Left (right) column shows network prediction accuracy using logits from randomly initialized (pretrained on ImageNet) networks. The logit vector is 2D. Second and third rows show accuracy as a function of the number of training data, using both normalized and unnormalized logits. The last row shows transfer results where a classifier trained on some networks is applied to the same networks but with different set of weights. Results are averaged over 5 runs.}
    \label{fig:BeesRandomPretrained}
\end{figure}

We trained the deep networks for binary classification of bees vs. ants using the {\bf BeesVsAnts} dataset from Kaggle\footnote{\url{https://www.kaggle.com/datasets/gauravduttakiit/ants-bees}}. This dataset has 244 training images and 153 testing images. Performance of deep networks over the test images are shown in the left panel of Fig.~\ref{fig:BeesRandomPretrained}. Deep nets are performing very well (above 80\%) and close to each other. The SVM classifiers, for network type prediction, are trained on 2196 training logits (244 per deep net; \ie 153 $\times$ 9), and are evaluated on 1377 testing logits (153 per deep net; \ie 153 $\times$ 9).
Notice that here the classification layer of deep nets consists of 2 neurons, and thus the logits are 2D vectors.

\textbf{Results using random and pretrained weights.} We fed the images to the deep nets with random weights or networks pretrained on ImageNet, and collected the logits. We then trained the SVM classifier on logits. 

Results are shown in Fig.~\ref{fig:BeesRandomPretrained}. With random weights, both SVM classifiers (linear and RBF) perform well above chance (1/9) using both normalized and unnormalized logits. With pretrained weights, both SVMs perform well in the unnormalized case. The RBF SVM performs consistently above chance in all cases on this dataset. Increasing the number of training data improves network prediction accuracy (except linear SVM in the unnormalized case over pretrained weights).

\begin{figure}[t]       
    \centering 
    \includegraphics[width=.47\linewidth]{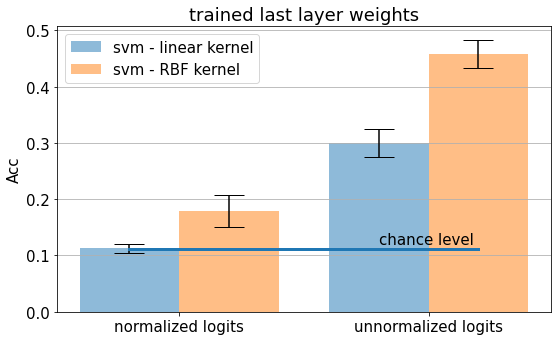}
    \includegraphics[width=.47\linewidth]{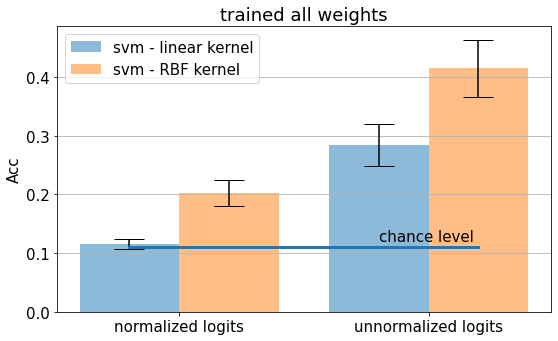}

    \includegraphics[width=.47\linewidth]{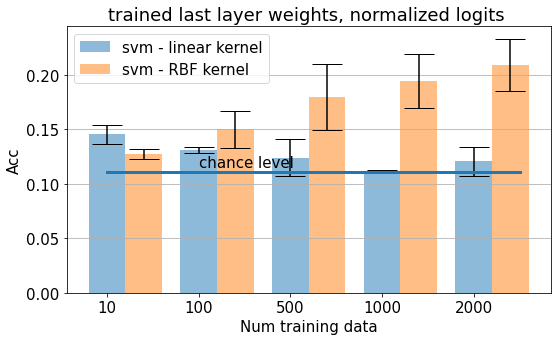}
    \includegraphics[width=.47\linewidth]{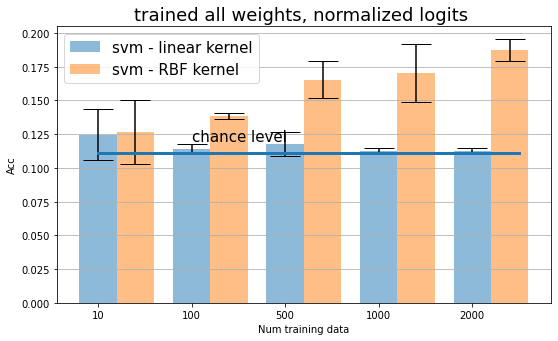}
    
    \includegraphics[width=.47\linewidth]{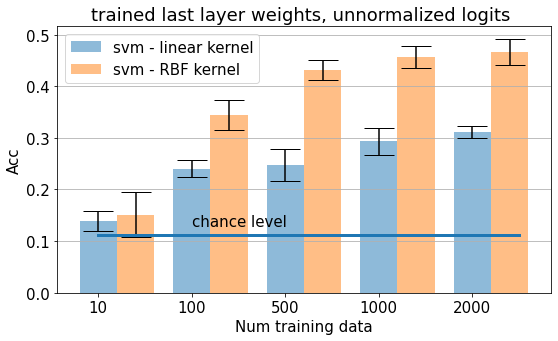}
    \includegraphics[width=.47\linewidth]{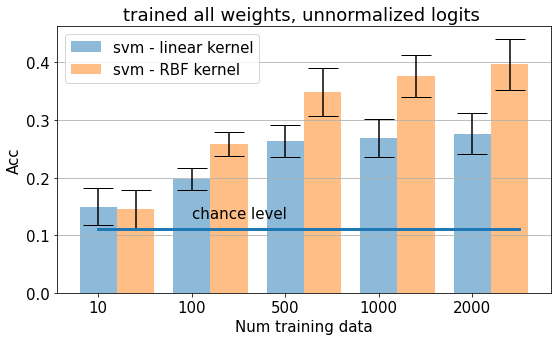}

    \includegraphics[width=.47\linewidth]{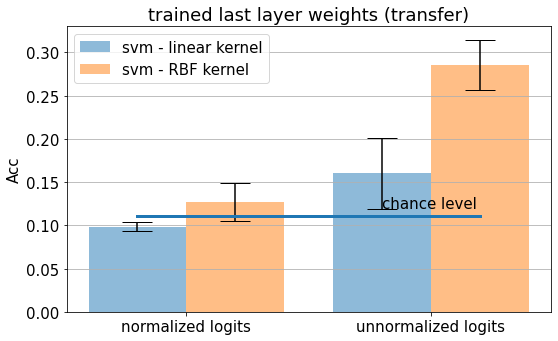}
    \includegraphics[width=.47\linewidth]{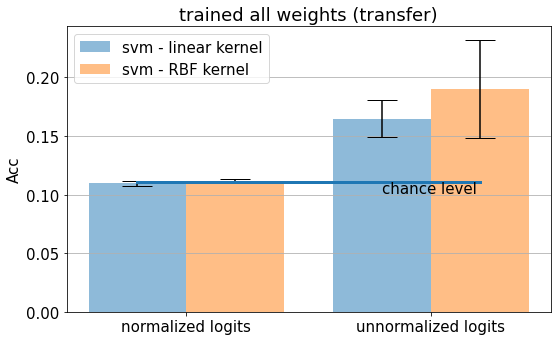}

    \caption{\textbf{Results over the BeesVsAnts dataset (cont'd)}. Left (right) column shows network prediction accuracy using logits from networks with their last layer (all layers) fine tuned (using networks pretrained on ImageNet). The logit vector is 2D. Second and third rows show accuracy as a function of the number of training data, using both normalized and unnormalized logits. The last row shows transfer results where a classifier trained on some networks is applied to the same networks but with different set of weights. Results are averaged over 5 runs.}    
    \label{fig:BeesTrained}
\end{figure}

\textbf{Results using trained weights:} Here we consider two cases: 1) logits from networks with their last layer fine-tuned, and 2) logits from networks with all weights fine-tuned. The networks are initialized with ImageNet weights. Results are shown in Fig.~\ref{fig:BeesTrained}. Here, both SVMs perform very well using unnormalized logits. Using normalized logits, however, only RBF SVM does better than chance. Increasing the number of training data improves network prediction accuracy (except linear SVM in the normalized case). Results are similar and consistent when training the last layer or all layers.

\textbf{Transfer Results.} According to the bottom panels in Figs.~\ref{fig:BeesRandomPretrained} and \ref{fig:BeesTrained}, a classifier trained on normalized logits does not transfer much, whereas in the unnormalized case RBF SVM shows a noticeable degree of generalization.

\subsection{Experiment II: CIFAR-10 classification}
Performance of the deep networks over the test set of CIFAR-10 dataset is shown in the right panel of Fig.~\ref{fig:BeesRandomPretrained}. Most of the networks perform above 60\% with Deit achieving around 80\%. SVM classifiers are trained over 8991 logits (999 per network; \ie 9 $\times$ 999) and tested over 8991 testing logits. Notice that here the classification layer of deep nets consists of 10 neurons, and thus the logits are 10D vectors.

\begin{figure}[t]       
    \centering 
    \includegraphics[width=.47\linewidth]{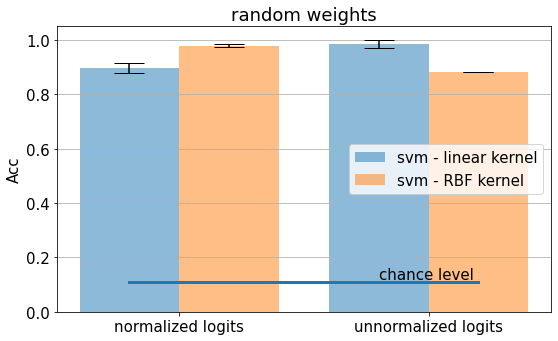}
    \includegraphics[width=.47\linewidth]{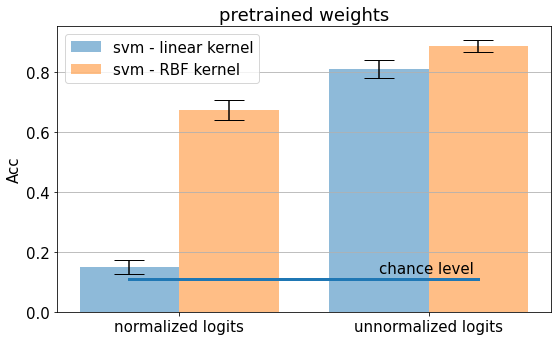}
    
    \includegraphics[width=.47\linewidth]{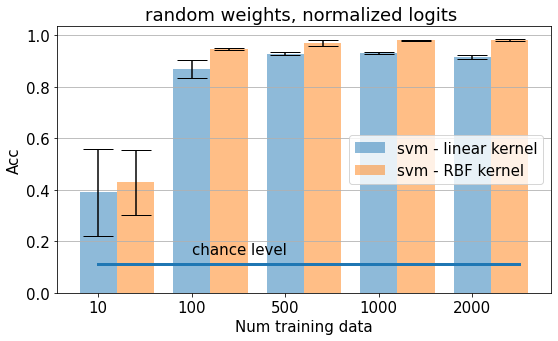}
    \includegraphics[width=.47\linewidth]{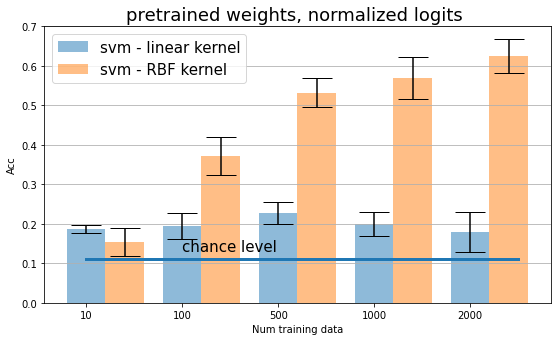}

    \includegraphics[width=.47\linewidth]{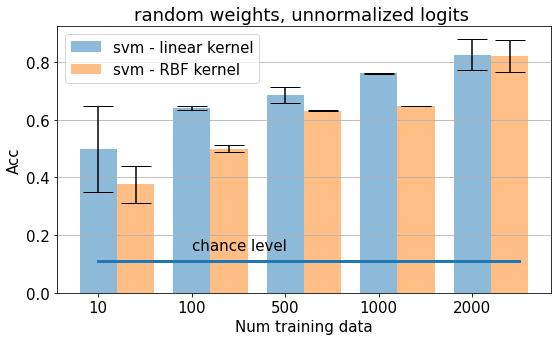}
    \includegraphics[width=.47\linewidth]{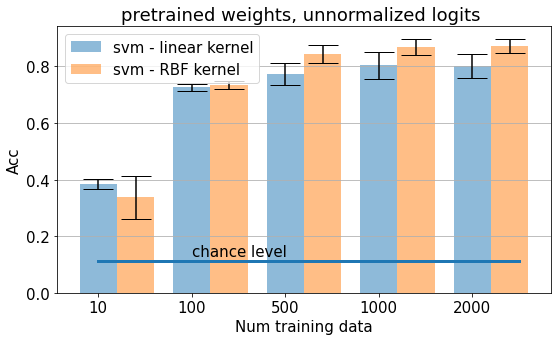}
    
    \includegraphics[width=.47\linewidth]{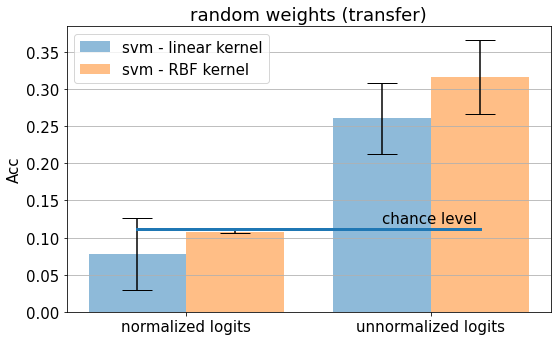}
    \includegraphics[width=.47\linewidth]{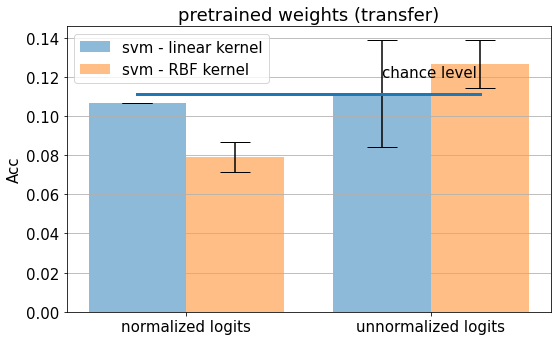}

    \caption{\textbf{Results over the CIFAR-10 dataset}. Left (right) column shows network prediction accuracy using logits from randomly initialized (pretrained on ImageNet) networks. The logit vector is 10D. Second and third rows show accuracy as a function of the number of training data, using both normalized and unnormalized logits. The last row shows transfer results where a classifier trained on some networks is applied to the same networks but with different set of weights. Results are averaged over 5 runs.}
    \label{fig:CIFARRandomPretrained}
\end{figure}

\textbf{Results using random and pretrained weights:} Consistent with results over the BeesVsAnts dataset, network prediction accuracy is very high here using logits derived from networks with random weights (Fig.~\ref{fig:CIFARRandomPretrained}). RBF SVM performs well using unnormalized logits from the pretrained networks.

\begin{figure}[t]       
    \centering 
    \includegraphics[width=.47\linewidth]{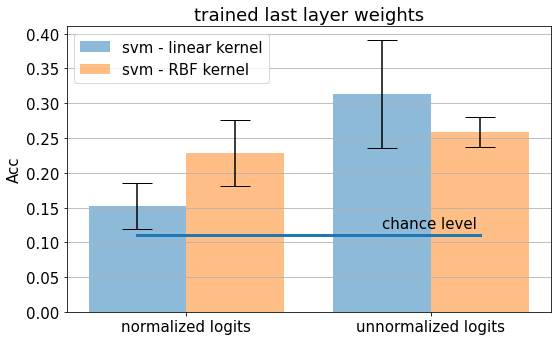} \\
    \includegraphics[width=.47\linewidth]{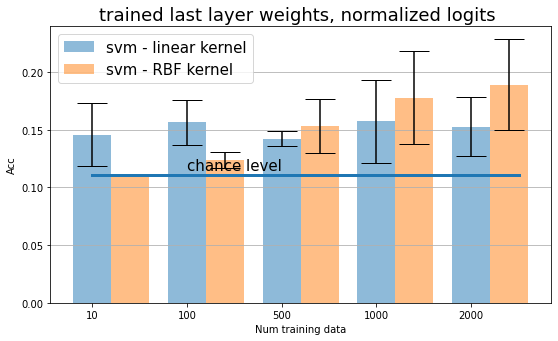}\\
    \includegraphics[width=.47\linewidth]{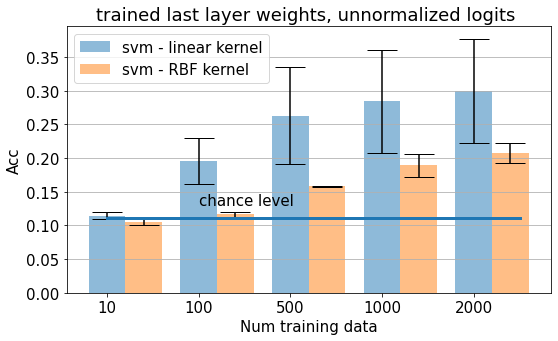}\\
    \includegraphics[width=.47\linewidth]{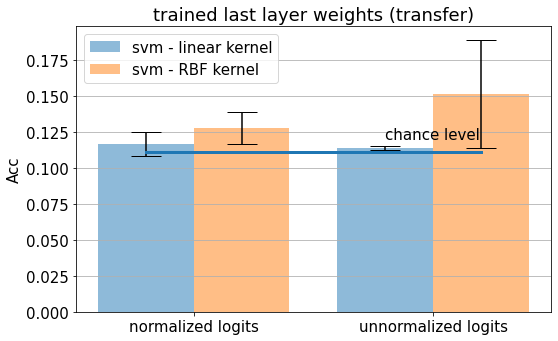}

    \caption{\textbf{Results over the CIFAR-10 dataset (cont'd)}. Network prediction accuracy using logits from networks with their last layer fine tuned (using networks pretrained on ImageNet). The logit vector is 10D. Second and third rows show accuracy as a function of the number of training data, using both normalized and unnormalized logits. The last row shows transfer results where a classifier trained on some networks is applied to the same networks but with different set of weights. Results are averaged over 5 runs.}    
    \label{fig:CIFARTrained}
\end{figure}

\textbf{Results using trained weights:} Due to limited computational resources, here we only fine-tune the weights of the last layer of deep networks (networks are initialized with ImageNet weights). Results are shown in Fig.~\ref{fig:CIFARTrained}. Consistent with corresponding results over the BeesVsAnts dataset, both SVMs perform significantly better than chance using unnormalized logits. RBF SVM performs significantly above chance using normalized logits over BeesVsAnt and CIFAR-10 datasets.

\textbf{Transfer Results.} Please see the last rows of Figs.~\ref{fig:CIFARRandomPretrained} and~\ref{fig:CIFARTrained}. Again, consistent with results over BeesVsAnts dataset, we do not observe a significant transfer.

\begin{figure}[t]       
    \centering 

    \includegraphics[width=.8\linewidth]{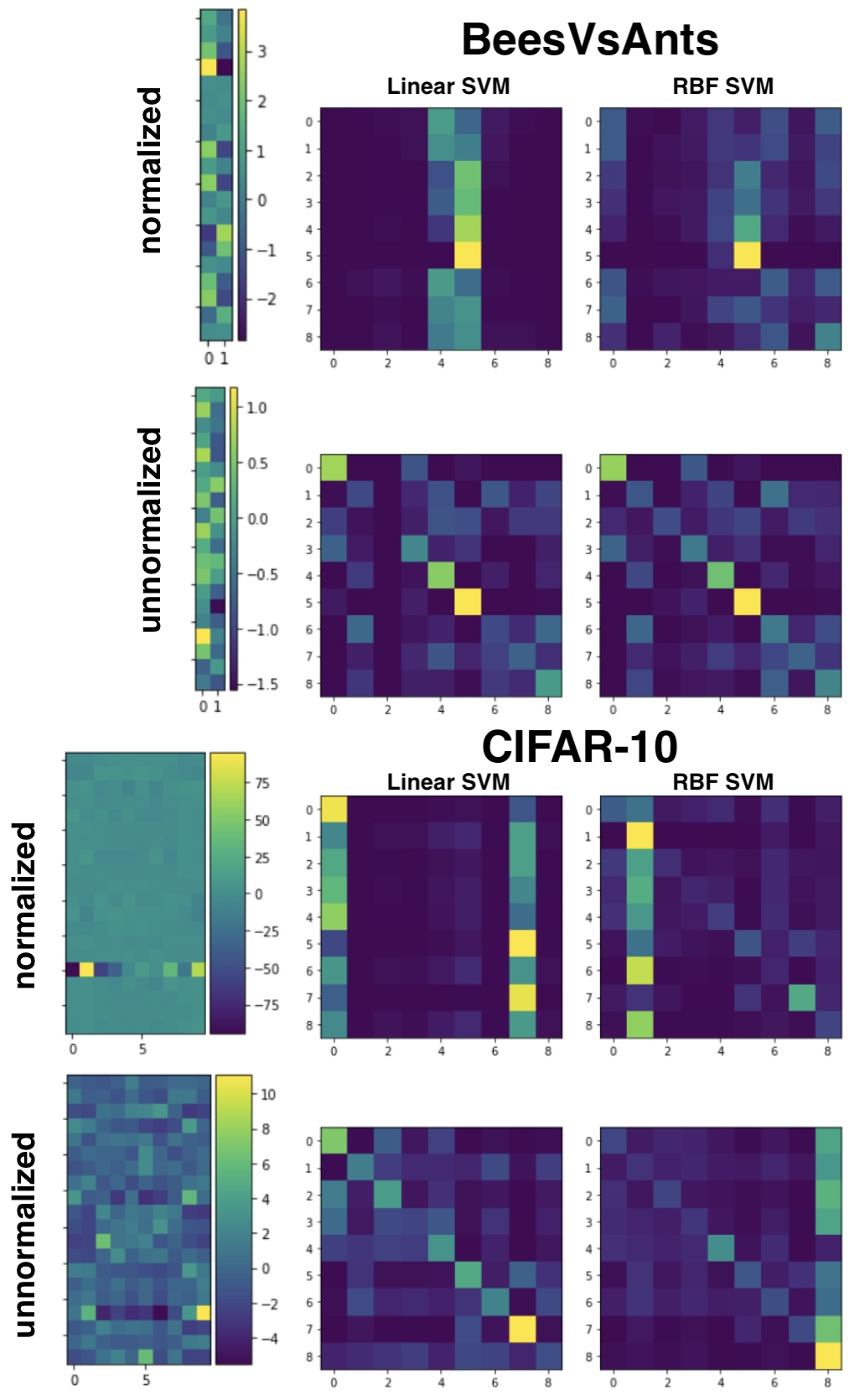}    
    
    \caption{\textbf{Visualization of logits and confusion matrices}. Twenty sample logit vectors (left column) along with sample confusion matrices of network classifiers over BeesVsAnts (top) and CIFAR-10 dataset using both normalized and unnormalized logit vectors. The row and column indices in the confusion matrices correspond to deep networks. Logits come from different deep networks. The logits on BeesVsAnts dataset are coming from a pretrained (on ImageNet) network, whereas the logits over CIFAR-10 are from trained networks (last layer).}
    \label{fig:samples}
\end{figure}

Fig.~\ref{fig:samples} shows sample normalized and unnormalized logit vectors, as well as confusion matrices of network classifiers.

\subsection{Experiment III: Classification using networks built for ImageNet}
In experiments above, we adjusted the classification layer of deep networks to be compatible with the dataset they were applied to. Here, we use the original deep nets constructed for ImageNet image classification. We feed images to these networks and record the 1000D logit vectors, and then train the SVM classifiers on these logits.  

\begin{figure}[t]       
    \centering 
    \includegraphics[width=.47\linewidth]{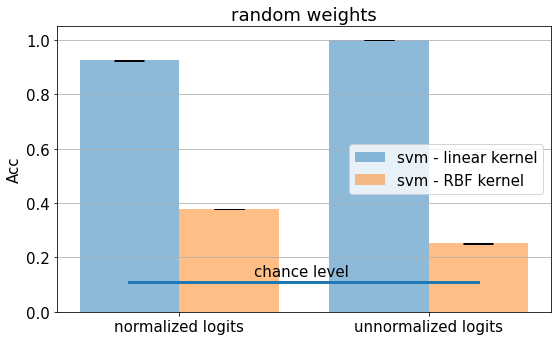}
    \includegraphics[width=.47\linewidth]{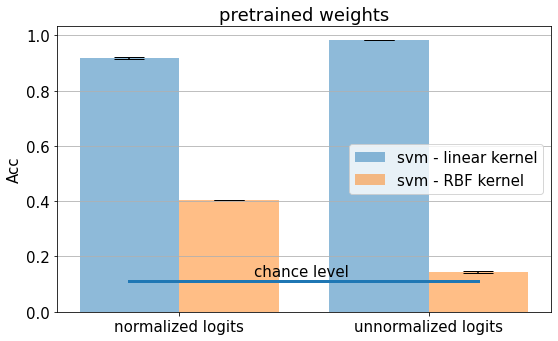}

    \includegraphics[width=.47\linewidth]{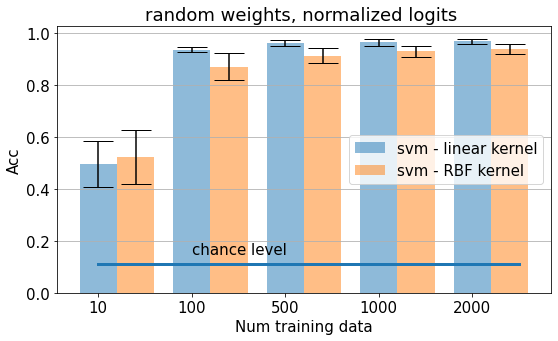}
    \includegraphics[width=.47\linewidth]{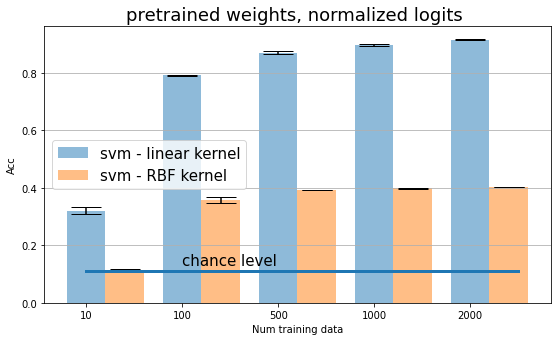}

    \includegraphics[width=.47\linewidth]{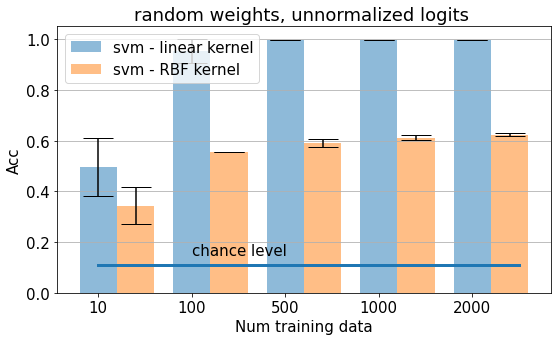}
    \includegraphics[width=.47\linewidth]{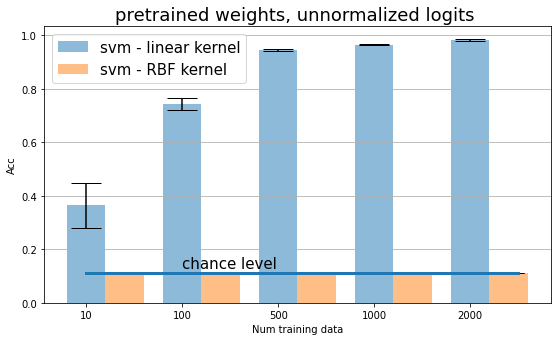}

    \caption{\textbf{Network prediction using ImageNet networks and feeding BeesVsAnts images.} 
    Left (right) column shows network prediction accuracy using logits from randomly initialized (pretrained on ImageNet) networks. The logit vector is 1000D. Second and third rows show accuracy as a function of the number of training data, using both normalized and unnormalized logits. Results are averaged over 5 runs.}
    \label{fig:ImageNetResBees}
\end{figure}

\begin{figure}[t]       
    \centering 
    \includegraphics[width=.47\linewidth]{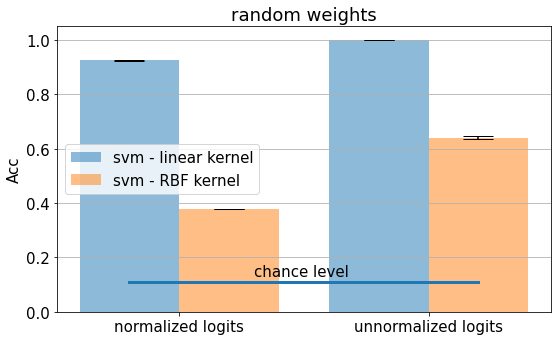}
    \includegraphics[width=.47\linewidth]{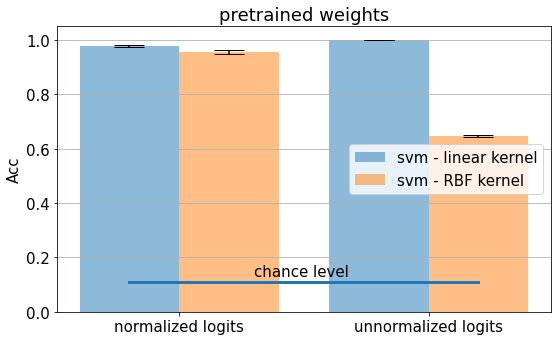}

    \includegraphics[width=.47\linewidth]{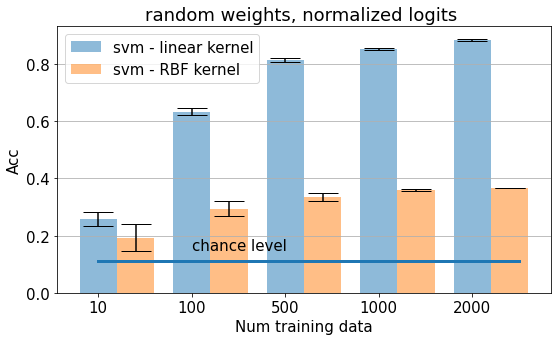}
    \includegraphics[width=.47\linewidth]{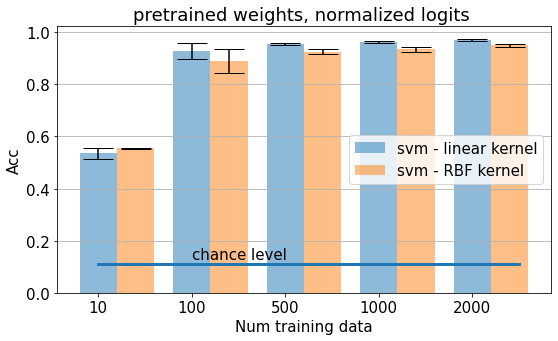}

    \includegraphics[width=.47\linewidth]{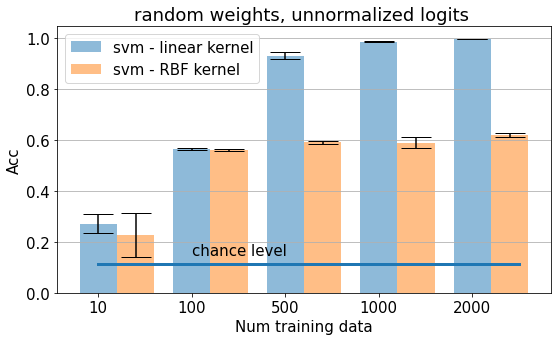}
    \includegraphics[width=.47\linewidth]{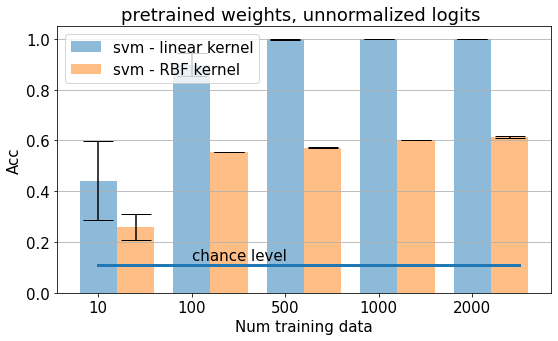}

    \caption{\textbf{Network prediction using ImageNet models and feeding CIFAR10 images.} 
    Left (right) column shows network prediction accuracy using logits from randomly initialized (pretrained on ImageNet) networks. The logit vector is 1000D. Second and third rows show accuracy as a function of the number of training data, using both normalized and unnormalized logits. Results are averaged over 5 runs.}
    \label{fig:ImageNetResCIFAR}
\end{figure}

\textbf{Results using random and pretrained weights.}
Results over BeesVsAnts and CIFAR-10 datasets are shown in Fig.~\ref{fig:ImageNetResBees} and Fig.\ref{fig:ImageNetResCIFAR}, respectively. Over both datasets, the linear SVM does pretty well using normalized and unnormalzied logits from random and pretrained networks. The RBF SVM does above chance, but lower than the linear SVM. This result suggests that network prediction might be possible even when classifiers are trained from logits in one domain and are tested on logits in another domain. This, however, needs further verification. One way to verify this, is to train the classifier with logits coming from images in domain A and to test it on logits coming from domain B (\eg training on BeesVsAnts and testing on CIFAR-10 using 1000D logit vectors).

\subsection{Experiment IV: MNIST classification}
\begin{figure}[t]       
\centering
    \includegraphics[width=.475\linewidth]{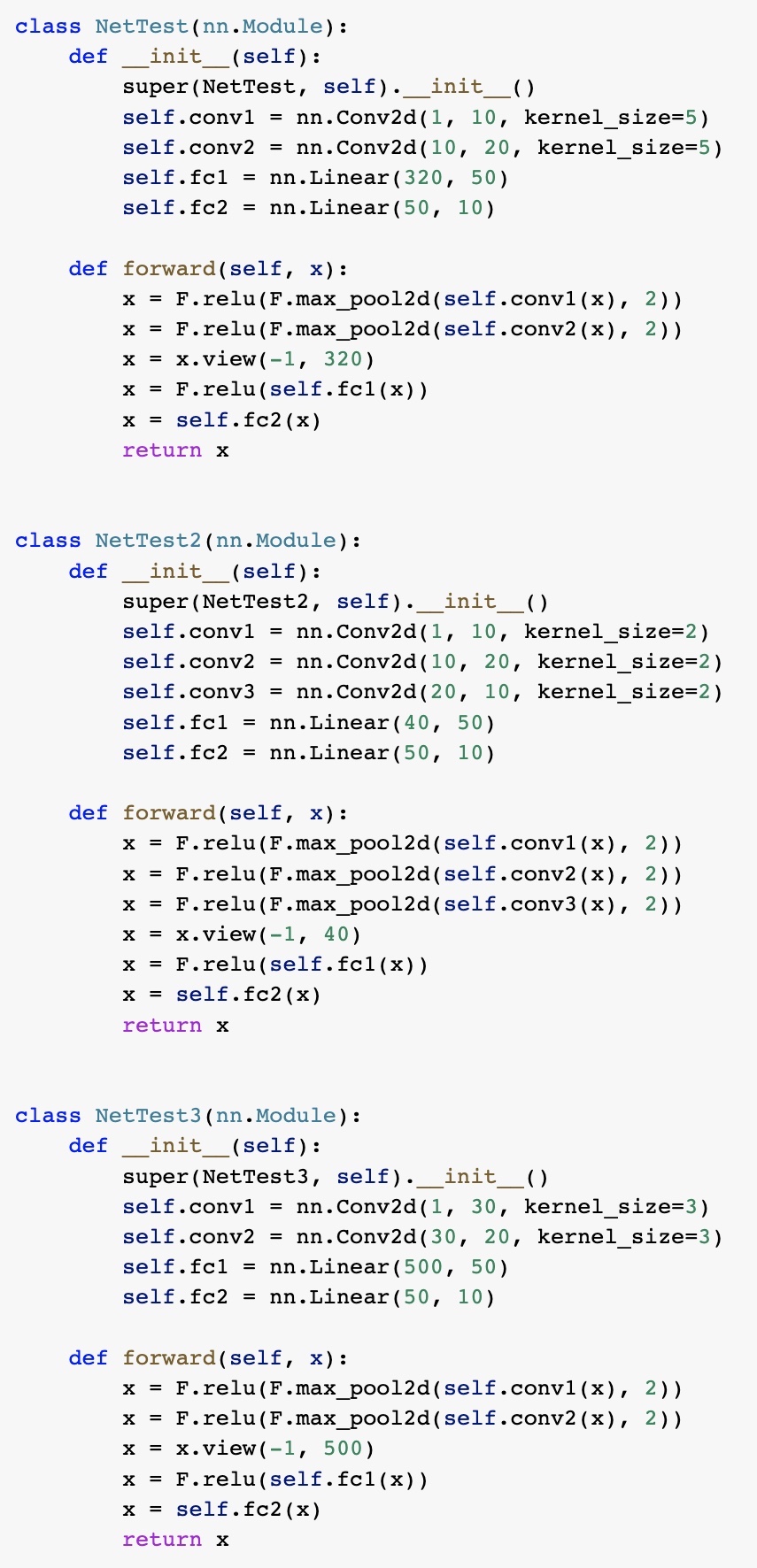}
    \includegraphics[width=.486\linewidth]{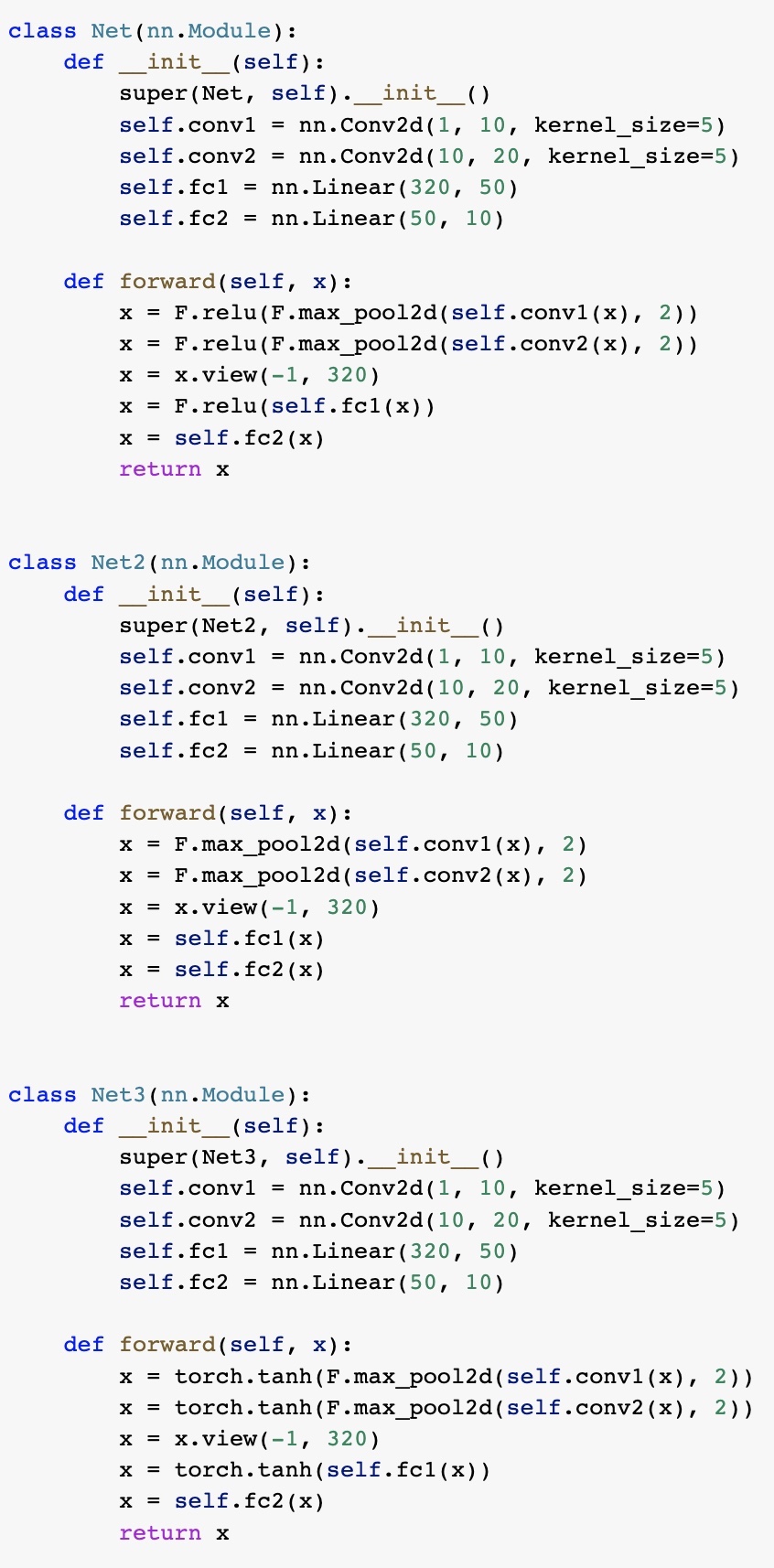}
    \caption{Left) CNN architectures used in the first MNIST experiment, Right) CNN architectures used in the second MNIST experiment.}
    \label{fig:mnistNets}
\end{figure}

So far we have shown that it is possible to predict the network type over heterogeneous networks, although some were very similar (\eg ResNet101 and ResNet152). Here, we consider prediction of homogeneous networks. We ask whether it is possible to predict the network type from modifications of the same network. The SVM classifiers here are trained over 8997 training logits and tested on 8997 logits. The logit vector is 10D. We perform the following two experiments:

\textbf{MNIST I.} Three CNNs with different numbers of layers and kernel sizes are constructed (left panel in Fig.~\ref{fig:mnistNets}). They have the same activation functions. 
The average performance of these CNNs for MNIST digit classification in order are 97.1\%, 86.5\%, and 96.7\%. 

\begin{figure}[t]       
    \centering 
    \includegraphics[width=.47\linewidth]{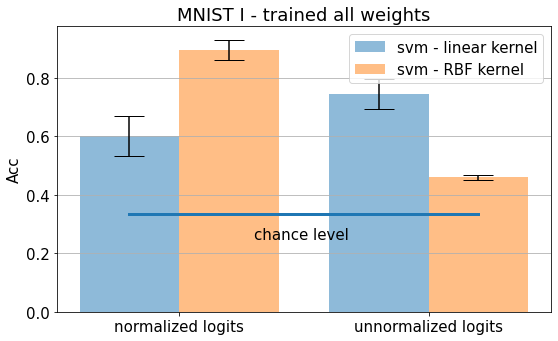}
    \includegraphics[width=.47\linewidth]{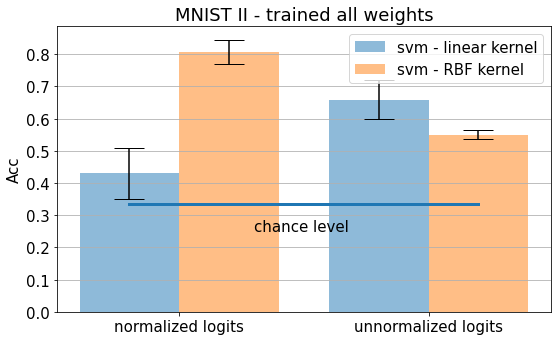}

    \includegraphics[width=.47\linewidth]{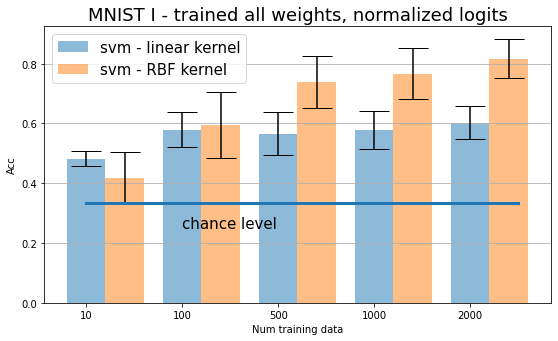}
    \includegraphics[width=.47\linewidth]{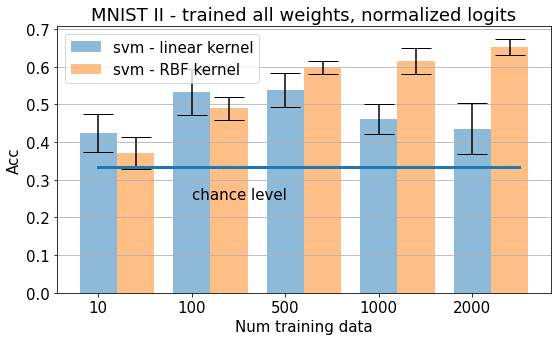}

    \includegraphics[width=.47\linewidth]{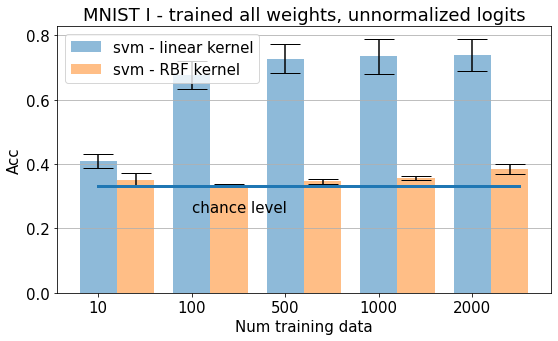}
    \includegraphics[width=.47\linewidth]{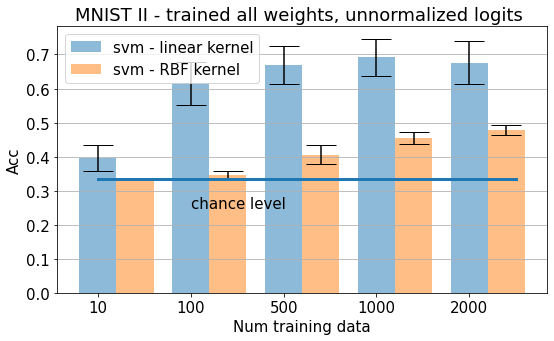}
    
    \caption{\textbf{Network prediction accuracy over MNIST dataset.} 
Left) network prediction accuracy over three CNNs with different number of layers and kernel sizes but with the same activation functions. Right) network prediction accuracy over three CNNs with the same number of layers and same kernel sizes but with different activation functions (ReLU, linear, and Tanh). See Fig.~\ref{fig:mnistNets} for definition of these CNNs. The logit vector is 10D. Second and third rows show accuracy as a function of the number of training data, using both normalized and unnormalized logits. Results are averaged over 5 runs.}
    \label{fig:mnist}
\end{figure}

We train 3-way SVMs to predict the network type from normalized or unnormalized logits. 
Results are shown in the left panel of Fig.~\ref{fig:mnist}. Although the network prediction accuracy now is lower than the above two datasets, it is still much better than the chance level (1/3).  


\textbf{MNIST II.} Three CNNs with the same number of layers (four layers) and same kernel sizes are constructed (right panel in Fig.~\ref{fig:mnistNets}). They have different activation functions (ReLU, linear, and Tanh). The average performance of these CNNs in order are 97.1\%, 97.1\%, and 95.4\%. Network prediction accuracy is shown in the right panel of Fig.~\ref{fig:mnist}. 


Results from both MNIST experiments show that it is still possible to separate the networks even when they are very similar to each other. This means that internal dynamics and small modifications inside a network cause a shift in distribution of logits.

\section{Discussion and Conclusion}

We showed that logits can be regarded as signatures of deep networks. In other words, the way deep neural networks encode uncertainty in their prediction is different and can be used to predict their type.

Our main findings are summarized below:
\begin{itemize}
 \item It is possible to predict the network type from logits, even when networks are very similar. Results are consistent across datasets and experiments.
    
    \item Unnormalized logits are better suited for network prediction, mainly because ranges of values are different across the networks, 
        
    \item On average, the SVM with RBF kernel works better than the SVM with linear kernel in predicting the network type,
    
    \item Increasing the number of training logits improves the accuracy.     
    
    \item Network prediction accuracy is higher using logits produced by networks with random weights than logits from pretrained or fine-tuned networks. This means that training the networks on the same dataset makes the logits more similar to each other, and hence less separable,

    \item We did not find a significant transfer of the learned classifier for network prediction when the same networks are initialized or trained differently. This, however, is not a big constraint for practical purposes, 
    
    \item Training only the last layer of the networks or all layers does not make a big difference in network prediction accuracy,
    
\end{itemize}

We foresee the following directions for future research in this area:
\begin{itemize}
\item We encourage researchers to explore the extent and scope of this phenomenon. Understanding the underlying reason why this phenomenon happens, in particular results using normalized logits, can provide new insights pertaining to how deep networks learn high dimensional representations. 

\item It might be possible to obtain better network prediction accuracy using additional tricks.
Including more features from networks such as weights from different layers may increase the network prediction accuracy. However, it does limit the applicability of the approach since such weights are not usually available at the inference time.

\item We suggest applying the proposed method to a larger number of network types, and to inspect for which images the network prediction is not possible and why.

\end{itemize}

{\small
\bibliographystyle{plain}
\bibliography{refs}
}

\end{document}